\documentclass{article}
\usepackage{PRIMEarxiv}

\usepackage[utf8]{inputenc} 
\usepackage[T1]{fontenc}    
\usepackage{hyperref}       
\usepackage{url}            
\usepackage{booktabs}       
\usepackage{amsfonts}       
\usepackage{nicefrac}       
\usepackage{microtype}      
\usepackage{lipsum}
\usepackage{fancyhdr}       
\usepackage{graphicx}       
\graphicspath{{media/}}     
\usepackage{amsmath}
\usepackage{amssymb}
\usepackage{cleveref}
\usepackage[official]{eurosym}
\usepackage{float}
\usepackage{array,tabularx}
\usepackage{booktabs} 
\usepackage{makeidx}
\usepackage{makecell}
\usepackage{xparse}
\usepackage{moredefs}
\usepackage{lips}
\usepackage{xcolor}

\newcommand\citep[1]{\cite{#1}}
\newcommand\citet[1]{\cite{#1}}

\title{Comparative Analysis of Evolutionary Algorithms for Energy-Aware Production Scheduling
}

\author{
  Sascha C Burmeister, Till N Rogalski, Guido Schryen \\
  Management Information Systems \\
  Paderborn University \\
  Paderborn, Germany\\
  sascha.burmeister@upb.de\\
}

\begin{document}
\maketitle

\begin{abstract}
The energy transition is driving rapid growth in renewable
energy generation, creating the need to balance energy supply and demand
with energy price awareness. One such approach for manufacturers
to balance their energy demand with available energy is energyaware
production planning. Through energy-aware production planning,
manufacturers can align their energy demand with dynamic grid conditions,
supporting renewable energy integration while benefiting from
lower prices and reduced emissions. Energy-aware production planning
can be modeled as a multi-criteria scheduling problem, where the objectives
extend beyond traditional metrics like makespan or required workers
to also include minimizing energy costs and emissions. Due to market
dynamics and the NP-hard multi-objective nature of the problem, evolutionary
algorithms are widely used for energy-aware scheduling. However,
existing research focuses on the design and analysis of single algorithms,
with limited comparisons between different approaches. In this study,
we adapt NSGA-III, HypE, and $\theta$-DEA as memetic metaheuristics for
energy-aware scheduling to minimize makespan, energy costs, emissions,
and the number of workers, within a real-time energy market context.
These adapted metaheuristics present different approaches for environmental selection.
In a comparative
analysis, we explore differences in solution efficiency and quality across
various scenarios which are based on benchmark instances from the literature
and real-world energy market data. Additionally, we estimate upper
bounds on the distance between objective values obtained with our
memetic metaheuristics and reference sets obtained via an exact solver.
\end{abstract}


\section{Introduction}
Energy-aware production planning incorporates environmental aspects and increased energy-awareness into production management and can be modeled as a multi-criteria job shop scheduling problem, where multiple jobs with varying processing times and job-dependent energy demand are scheduled on a machine \citep{baensch2021energy}.
Its objectives extend beyond traditional metrics like makespan or required workers to also include minimizing energy costs and emissions.
This paper focuses on the Flexible Job Shop Scheduling Problem (FJSP), in which each job consists of sequential operations that can be performed on one or more machines of a given set.

Given the NP-hard complexity and multi-criteria character, metaheuristic approaches are frequently preferred for their ability to adapt quickly to fluctuations of a dynamic market \citep{gao2020review}.
Multi-objective evolutionary algorithms (MOEAs) are population-based algorithms that aim to find a set of solutions close to the Pareto front while achieving a diverse, well-distributed coverage.
Based on their search strategies, they can be categorized into three main design paradigms: (1) dominance-based, (2) indicator-based, and (3) decomposition-based approaches \citep{bezerra2018large}.
MOEAs have been applied in the domain of energy-aware production planning with real-time pricing, with tailored adaptations to address specific problem requirements.
\cite{gong2018new} incorporates energy consumption alongside makespan and human factor indicators in their analysis, using hybrid GA and NSGA-II as dominance-based algorithms.
\cite{chen2022energy} implements decomposition-based MOEA/D \citep{zhang2007moead} to concurrently optimize makespan and energy costs, while \cite{dong2022green} integrates a swarm-based algorithm with NSGA-III to include emissions minimization as an additional objective.
Similarly, \cite{burmeister2025two} apply a memetic NSGA-III to quantify potential reductions in energy costs and emissions.
We identify the research gap that existing studies on energy-aware scheduling primarily examine individual algorithms, while comparative analyses in multiple MOEAs remain limited.

This work addresses this gap by comparing multiple memetic MOEAs for an energy-aware FJSP.
From each paradigm, we select one representative and enhance it with a memetic approach to improve problem adaptation.
We select decomposition-based NSGA-III for its widespread adoption and proven success in prior research \citep{jain2013evolutionary}.
We select decomposition-based $\theta$-DEA as it also works with dominance relations to enhance the convergence behavior of NSGA-III by leveraging the MOEA/D fitness evaluation scheme \citep{yuan2015new}.
We selected indicator-based HypE, as it showed strong performance in many-objective problems, but lacks application to energy-aware FJSPs \citep{bader2011hype}.
Our research question is:
How do the MOEAs considered vary in terms of solution efficiency and quality across different scenarios?
Our contributions include (1) formulating an energy-aware FJSP, (2) developing memetic NSGA-III, $\theta$-DEA, and HypE, and (3) conducting computational experiments to evaluate the algorithms.

This paper is structured as follows:
Sect. \ref{sec:model} defines the problem.
Sect. \ref{sec:methodology} details the algorithms and their memetic adaptation.
Sect. \ref{sec:compex-settings} and \ref{sec:compex-results} present the computational experiments, and
Sect. \ref{sec:conclusion} discusses limitations and future research.

\section{Energy-Aware Flexible Job Shop Scheduling Problem}\label{sec:model}
In this section, we present the mathematical optimization model for the energy-aware FJSP with the objectives of minimizing makespan, energy cost, emissions and the number of workers.
We extend the model of \cite{burmeister2025memetic} and enhance its complexity by adding the total number of workers required as an additional objective, informed by discussions with industry practitioners.
Table \ref{tab:notation} presents the notation for the mathematical model.
The set $J$ represents the jobs to be processed, each consisting of $v_i$ operations $O_i=\{(i,1),...,(i,v_i)\}$ that must be processed sequentially. 
Each operation is attributed with a machine-dependent processing time and an energy requirement.

\begin{table}[t]
	\centering
	\caption{Notation for the mathematical formulation}\label{tab:notation}
	\begin{tabular}{lp{5cm}|lp{4.7cm}}
		\hline
        \bf Term & \bf Description & \bf Term & \bf Description\\ 
        \hline
		\multicolumn{2}{l|}{\textit{Sets}} & \multicolumn{2}{l}{\textit{Variables}}\\
		$J$ & Jobs, $i \in J$ &
            $c^{max}$ & Maximum makespan\\
		$O$ & Operations, $O=\bigcup_{i\in J} O_i$, &
            $p^{sum}$ & Sum of all energy cost\\
        & $O_i=\{(i,1),...,(i,\nu_i)\}$ &
            $e^{sum}$ & Sum of all emissions\\
		$M$ & Machines, $k\in M$&
            $w^{max}$ & Maximum workers needed \\
		$T$ & Time steps, $t\in T$&
            $w_t$ & Workers needed at time~$t$\\
        \multicolumn{2}{l|}{\textit{Parameters}} & 
            $p_{ijkt}$ & Binary; 1 iff $(i,j)$ starts on $k$\\
        $\omega_{ij}$ & Worker demand for processing $(i,j)$&
            & at time~$t$\\
        $\tau_{ijk}$ & Processing time of $(i,j)$ on $k$&\\
		\hline
	\end{tabular}
\end{table}

The objective function (\ref{eq:obj}) aims to minimize four key variables: the makespan $c^{max}$, the sum of energy cost $p^{sum}$, the sum of emissions $e^{sum}$, and the number of workers needed $w^{max}$.
Constr. (2) - (14) are described in detail in \citep{burmeister2025memetic}.
They determine the makespan across all operations and machines, compute cumulative energy costs and emissions, ensure that each operation is assigned to exactly one machine, are processed sequentially and do not overlap on the same machine.

To integrate the number of required workers into the model, we add Constr. (\ref{eq:workforce-sum}) to sum the respective demand for workers for each time step $t$.
The variable $p_{ijkt'}$ indicates whether the operation $(i,j)$ starts on machine $k$ at time step $t'$.
The sum of $p_{ijkt'}$ over $t'\in[t-\tau_{ijk},t]$ is equal to one if operation $(i,j)$ is processed at time step $t$, so that $\omega_{ij}$ workers are required to process it.
Constr. (\ref{eq:workforce-max}) sets $w^{max}$ at the highest total workforce requirement $w_t$ in all time steps $t$.
The model comprises binary and continuous decision variables and combines linear and convex constraints, classifying it as a Mixed-Integer Linear Program (MILP).

\begin{align}
	\text{min } &(c^{max}, p^{sum}, e^{sum}, w^{max}) 	& \label{eq:obj} \\
	\text{s.t. }&(2) - (14) & \text{from \citep{burmeister2025memetic}}	\nonumber \setcounter{equation}{14}\\
    &w_t=\sum_{i,j}\omega_{ij}\sum_{k,t'=t-\tau_{ijk}}^{t}p_{ijkt'} & \forall t\label{eq:workforce-sum}\\
    &w^{max} \geq w_t                           & \forall t\label{eq:workforce-max}\\
	&w^{max}, w_t \in \mathbb{R}_+ ,  & \forall t
\end{align}

\section{Methodology}\label{sec:methodology}

\subsection{Representation} \label{ssec:representation} 
\begin{figure}[h]
    \centering
    \includegraphics[width=\textwidth]{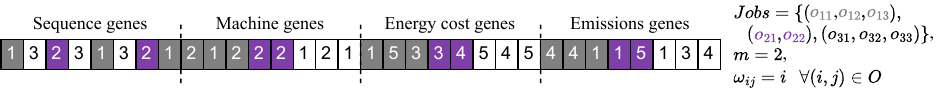}
    \caption{Sample genotype \citep{burmeister2025memetic}}
    \label{fig:genotype}
\end{figure}

We adopt a decoder-based representation.
For the encoded solution, we utilize the quadripartite gene string approach of \cite{burmeister2025memetic}, see \Cref{fig:genotype}.
The first gene string represents the sequence of operations, listing all jobs in the order in which their operations are scheduled.
The second gene string specifies the machine assignment for each operation.
The third and fourth gene strings correspond to energy costs and emissions, respectively, indicating the maximum acceptable values at which an operation is scheduled.

Figure 2 illustrates the phenotype in the form of a Gantt chart.
The right y-axis displays the energy cost values as a dashed line and the emissions values as a dotted line.
The sequence genes dictate that the first operation from job 1 must be assigned, and according to the machine genes, it is allocated to machine 2.
During scheduling, the earliest feasible time slot is selected where the specified maximum permissible values are satisfied, e.g. 1 monetary unit for energy costs and 4 units for emissions for operation $(1,1)$.

In this work, we add worker requirements for each time slot to the phenotype.
In the given example, an operation's worker demand is defined as $\omega_{ij}=i$.
The maximum worker demand $w^{max}$ is 5 at time step 2 when operations from jobs 2 and 3 are processed simultaneously.

\begin{figure}[h]
    \centering    
    \includegraphics[width=\textwidth]{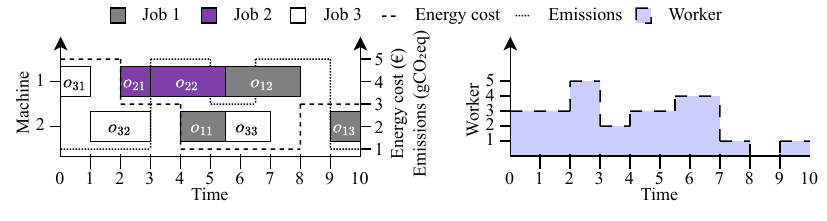}
    \caption{Sample phenotype for \Cref{fig:genotype}}
    \label{fig:phenotype}
\end{figure}
\clearpage
\subsection{Evolutionary Algorithms}\label{ssec:algorithms}
For this study, we select NSGA-III, $\theta$-DEA, and HypE as algorithms following different paradigms.
All algorithms are MOEAs, inherently population-based, and employ interchangeable operators for recombination and mutation.

\paragraph{NSGA-III}\label{ssec:nsga3} 
The NSGA-III divides its population into fronts using non-dominated sorting.
When NSGA-III selects a subset of individuals from a front $P_t$ in generation $t$ for the next generation $P_{t+1}$, it employs reference points in the objective space and clusters individuals based on their shortest perpendicular distance to these reference points.
Individuals are then added to $P_{t+1}$ in a manner that promotes an even distribution across the reference points until all available slots are filled.
For further details, see \citep{jain2013evolutionary}.

\paragraph{$\theta$-DEA}\label{ssec:theta-dea} 
The $\theta$-DEA defines clusters $C_j, j\in\{1,...,N\}$.
Function $\tilde{f}(x)$ represents the normalized solution vector of a solution $x$, $L$ is a line passing through the origin in a given direction of a reference vector $\lambda_j$, and $u$ is the projection of $\tilde{f}(x)$ onto $L$.
The evaluation metric $\mathcal{F}_j(x)=d_{j,1}(x)+\theta d_{j,2}(x)$ combines the distance $d_{j,1}(x)$ from the origin to $u$ and the perpendicular distance $d_{j,2}(x)$ between $\tilde{f}(x)$ and $L$, with the latter weighted by a predefined penalty parameter $\theta$.
Set $C_j$ comprises all individuals with minimum $d_{j,2}(x)$.
A classification into fronts is achieved via non-dominated sorting, employing $\theta$-dominance instead of Pareto dominance.
A solution $x$ $\theta$-dominates another solution $y$, iff $x\in C_j$, $y\in C_j$, and $\mathcal{F}_j(x)<\mathcal{F}_j(y)$.
The $\theta$-DEA aims to preserve diversity using non-dominated sorting while improving convergence through the use of $\theta$-dominance relation.
For further details, see \citep{yuan2015new}.

\paragraph{HypE}\label{ssec:hype} 
The HypE assigns fitness values using the hypervolume indicator, estimated via Monte Carlo simulation to manage computational complexity.
For an objective space $Z\in\mathbb{R}^n$, it uses a sampling set $S:=\{(z_1,...,z_n)\in Z\vert \forall1\leq i\leq n:l_i\leq z_i\leq u_i\}$.
The lower bound $l_i:\min_{a\in P}f_i(a)$ is the minimum fitness of all individuals $a$ in the population $P$.
The upper bound $u_i:=\max_{(r_1,...,r_n)\in R}r_i$ is given by a reference set $R\subset Z$ of non-dominated objective vectors.
Sampling $M$ vectors uniformly at random from $S$, the hypervolume is approximated as the product of the sampling box volume $V$ and the proportion of sampled points dominating $R$ and dominated by $P$.
For selection, HypE prioritizes solutions that maximize hypervolume coverage.
For further details, see \citep{bader2011hype}.

\subsection{Memetic framework}\label{ssec:memetic}
\begin{figure}[h]
    \includegraphics[width=\textwidth]{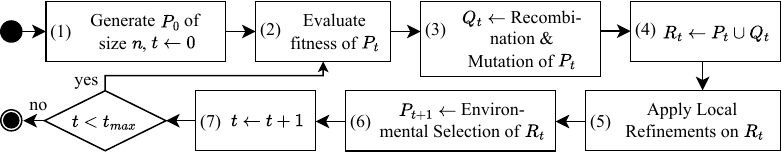}
    \caption{Algorithm framework}
    \label{fig:framework}
\end{figure}

To extend the algorithms to memetic NSGA-III (MNSGA-III), memetic $\theta$-DEA ($\theta$-DEA), or memetic HypE (MHypE), we integrate two local refinement methods.
\Cref{fig:framework} shows the framework that begins with (1) generating the initial population $P_0$ and (2) evaluating its fitness.
(3) Recombination and mutation are then applied using operators from \citet{burmeister2025memetic}, followed by (4) merging of the parent and offspring populations.
(5) Selected individuals undergo local refinement through a greedy method from \citet{burmeister2025memetic}, optimizing energy costs without increasing makespan, and an adapted approach for workforce allocation. 
After algorithm-specific environmental selection (6, see Sect. \ref{ssec:algorithms}), the next generation $P_{t+1}$ is determined.
This process repeats until the desired number of generations is reached.

\section{Experimental Settings}\label{sec:compex-settings}
We use the benchmark set of \cite{brandimarte1993routing}, enhanced with energy demands, prices, and emissions, as in \citep{burmeister2025memetic}, and add worker demands $\omega_{ij}\in[1,4]$ for each operation $(i,j)$, with the assignment: $\omega_{ij}=j\text{ mod }(i\text{ mod }4+1)+1$.

For parameter configuration, all algorithms use the settings of~\cite{burmeister2025memetic}.
Deviating from this, we increase the population size to 1000 and reduce the mutation rate to 0.1, since the algorithms generate large non-dominant fronts.
M$\theta$-DEA applies $\theta=5$ as in \cite{yuan2015new}.
MHypE sets the sampling size equal to the population size.

For evaluating solution quality, we use the $\varepsilon$-constraint method to compute a reference set $R$ of weakly Pareto optimal points.
First, we constrain the makespan to four equally distributed values between best known value \citep{li2016effective} and worst value found.
For each of these makespan values, we iteratively impose constraints on the remaining three objectives:
One of the remaining objectives (energy cost, emissions, or number of workers) is restricted using four equally distributed values within its observed range.
For each combination of makespan and the first constrained objective, another objective from the remaining two is also constrained using four equally spaced values.
Finally, a third objective is constrained using four equally distributed values while the fourth objective remains unconstrained and is minimized.
In total, this leads to $3$ (objective combinations) $\times 4$ ($\varepsilon$-values per objective) $\times 4\times  4 = 192$ solutions per instance.

The experiments run on Red Hat Enterprise Linux 8.5 (Oopta) with an Intel Xeon Gold 6148 CPU (20×2.4GHz) and 190 GB RAM.
Due to randomness, each algorithm is run 10 times.
For $\varepsilon$-constraint method, we use Gurobi v11 with a runtime limit of 12 hours.

\section{Results}\label{sec:compex-results}

\begin{figure}[h]
    \centering    
    \includegraphics[width=\textwidth]{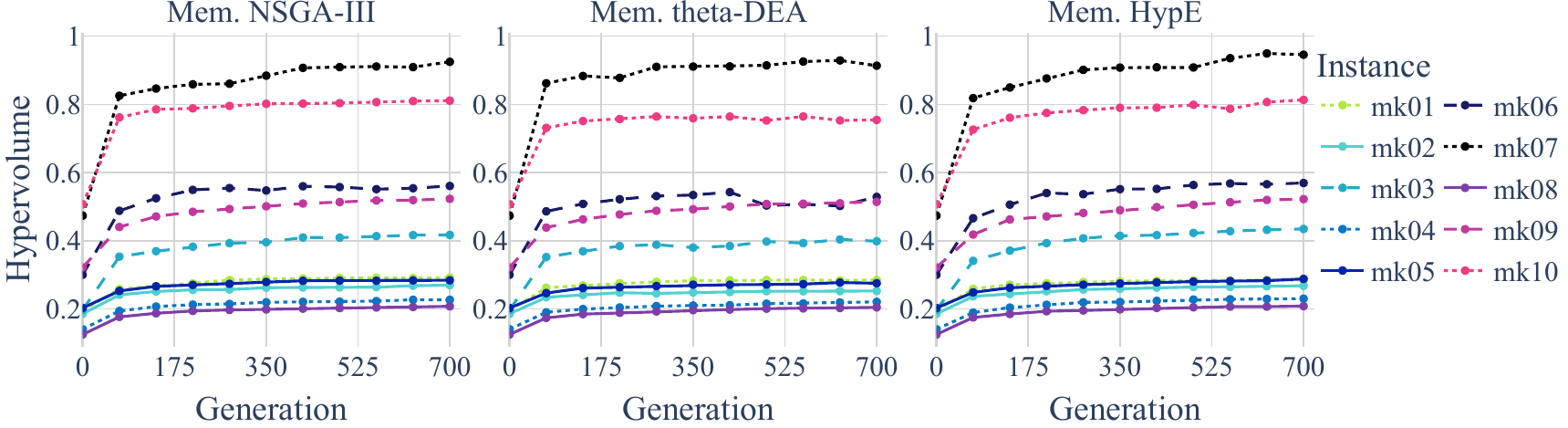}
    \caption{Growth in hypervolume per instance over generations}
    \label{fig:hypervolume}
\end{figure}

\Cref{fig:hypervolume} shows the relative improvement in hypervolume for the three algorithms over 700 generations.
The hypervolume indicator measures the volume in the objective space that is dominated by a given non-dominated front \citep{bader2011hype}.
For each instance, we normalize the solutions to a [0,1] scale using the best and worst objective values per dimension across all algorithms and the exact method.
As the utopian point may be infeasible, we analyze hypervolume progression over generations rather than a single value. 
The most significant gains occur early, with increases of $24.7$–$79.0$\% (MNSGA-III), $21.6$–$82.0$\% (M$\theta$-DEA), and $22.6$–$72.8$\% (MHypE) from generation 0 to 70.
Over 700 generations, hypervolume grows by $40.0$–$111.2$\% (MNSGA-III), $35.8$–$101.9$\% (M$\theta$-DEA), and $42.1$–$120.1$\% (MHypE).

\Cref{tab:results} presents metrics for evaluation.
The column $\vert R\vert$ shows the number of reference points, while gap\textsubscript{max} indicates the maximum optimality gap of points in $R$.
Spacing assesses the uniformity of point distribution on the heuristic front.
Modified Inverted Generational Distance (IGD$^+$) measures the average distance between non-dominated reference points and their nearest neighbors on the heuristic front, while a Modified Generational Distance (GD$^+$) measures the distance between non-dominated points from the heuristic and points in $R$ \citep{ishibuchi2015modified}.
HV$_R-$HV$_{F0}$ quantifies the hypervolume difference between $R$ and the heuristic front, representing the space not dominated by the best front.

\begin{table}[H]
	\centering
    \addtolength{\leftskip} {-2cm}
    \addtolength{\rightskip}{-2cm}
	\caption{Metrices}\label{tab:results}
    \fontsize{8}{10}\selectfont
	\begin{tabular}{lcrc|ccc|ccc|ccc|ccc}    
    \hline
    Inst. &\multicolumn{3}{c}{$R$} & \multicolumn{3}{c}{IGD$^+$} & \multicolumn{3}{c}{GD$^+$} & \multicolumn{3}{c}{HV$_{R}$-HV$_{F0}$}  & \multicolumn{3}{c}{Spacing} \\
    & $\vert R\vert$ & Gap\textsubscript{max} & HV$_R$ & N & $\theta$ & H & N & $\theta$ & H & N & $\theta$ & H & N & $\theta$ & H \\
    \hline
    mk01 & 104 & 65.6 & 0.487 & 0.227 & 0.224 & 0.228 & 0.113 & 0.085 & 0.134 & 0.195 & 0.202 & 0.202 & 0.021 & 0.030 & 0.022 \\
    mk02 & 97 & 108.5 & 0.461 & 0.256 & 0.269 & 0.259 & 0.170 & 0.095 & 0.180 & 0.191 & 0.209 & 0.193 & 0.020 & 0.026 & 0.020 \\
    mk03 & 84 & 5502.5 & 0.662 & 0.207 & 0.210 & 0.196 & 0.029 & 0.021 & 0.032 & 0.245 & 0.263 & 0.227 & 0.017 & 0.027 & 0.016 \\
    mk04 & 93 & 176.5 & 0.472 & 0.287 & 0.287 & 0.285 & 0.045 & 0.033 & 0.039 & 0.245 & 0.251 & 0.242 & 0.022 & 0.032 & 0.024 \\
    mk05 & 79 & 180.4 & 0.454 & 0.247 & 0.252 & 0.255 & 0.061 & 0.048 & 0.059 & 0.170 & 0.178 & 0.166 & 0.022 & 0.029 & 0.019 \\
    mk06 & 84 & 1059.8 & 0.708 & 0.146 & 0.165 & 0.150 & 0.015 & 0.010 & 0.017 & 0.147 & 0.179 & 0.138 & 0.015 & 0.106 & 0.012 \\
    mk07 & 82 & 142.2 & 0.857 & 0.075 & 0.071 & 0.068 & 0.019 & 0.006 & 0.021 & 0.173 & 0.181 & 0.158 & 0.006 & 0.010 & 0.006 \\
    mk08 & 47 & 214.0 & 0.451 & 0.328 & 0.324 & 0.325 & 0.086 & 0.056 & 0.102 & 0.244 & 0.247 & 0.243 & 0.026 & 0.033 & 0.025 \\
    mk09 & 0 & - & 0.000 & - & - & - & - & - & - & -0.523 & -0.514 & -0.522 & 0.023 & 0.040 & 0.025 \\
    mk10 & 0 & - & 0.000 & - & - & - & - & - & - & -0.811 & -0.754 & -0.813 & 0.014 & 0.158 & 0.078 \\
    \hline
    \multicolumn{16}{l}{N: MNSGA-III, \quad $\theta$: M$\theta$-DEA, \quad H: MHypE}
\end{tabular}
\end{table}

The results for $R$ highlight the problem's complexity, as no points were found for mk09 and mk10, while others show high optimality gaps.
Consequently, some heuristically computed points may dominate points in $R$.
To address this, we present both IGD$^+$ and GD$^+$.
IGD$^+$ shows that the points in $R$ remain 0.068 to 0.328 away from the heuristic points, with a maximum possible distance of 2 in a four-dimensional space.
GD$^+$ reveals that all heuristics identify solutions dominating parts of $R$, with distances ranging from 0.010 to 0.18.
Subtracting HV$_{F0}$ from HV$_R$ shows that the largest difference is 0.263 (mk03, M$\theta$-DEA), while the smallest positive difference is 0.138 (mk06, MHypE). 
MNSGA-III has the smallest difference in 3/10 cases, while MHypE performs best in 7/10.
Spacing values remain below 0.03 for almost all algorithms and instances, reflecting evenly distributed fronts.
MNSGA-III achieves the lowest spacing in 6/10 cases, while MHypE performs best in 6/10, with ties for mk02 and mk07.

\section{Conclusion}\label{sec:conclusion}
We adapt memetic NSGA-III, $\theta$-DEA, and HypE to the energy-aware FJSP and perform a comparative analysis.
All algorithms provide at least an acceptable approximation to the reference set. 
Notably, all algorithms were able to dominate parts of the reference set, while exact methods failed to find optimal solutions within the given runtime due to the problem's complexity. 
The computed non-dominated fronts exhibit good spacing, with MHypE performing best in most instances in terms of both solution quality and spacing.
The results indicate that HypE's search strategy is particularly well suited for the problem instances at hand.
Future research should explore real-world instances and incorporate parameter tuning to ensure that suboptimal results are not due to inappropriate configuration of parameters.

\section*{Acknowledgments}
The authors gratefully acknowledge the financial support provided by the state of North Rhine-Westphalia, Germany, as part of the \textit{progres.nrw} program area, in the framework of Re$^2$Pli (project number EFO 0127A) and the funding of this project by computing time provided by the Paderborn Center for Parallel Computing (PC$^2$).


\end{document}